# Explainable Probabilistic Machine Learning for Predicting Drilling Fluid Loss of Circulation in Marun Oil Field

Seshu Kumar Damarla
Department of Chemical and Materials Engineering
Univeristy of Alberta
Edmonton, Canada
damarla@ualberta.ca

Xiuli Zhu
Department of Optical-Electrical and Computer Engineering
University of Shanghai for Science and Technology
Shanghai, P. R. China
xiulizhu@usst.edu.cn

*Abstract*—Lost circulation remains a major and costly challenge in drilling operations, often resulting in wellbore instability, stuck pipe, and extended non-productive time. Accurate prediction of fluid loss is therefore essential for improving drilling safety and efficiency. This study presents a probabilistic machine learning framework based on Gaussian Process Regression (GPR) for predicting drilling fluid loss in complex formations. The GPR model captures nonlinear dependencies among drilling parameters while quantifying predictive uncertainty, offering enhanced reliability for high-risk decision-making. Model hyperparameters are optimized using the Limited-memory Broyden–Fletcher–Goldfarb–Shanno (L-BFGS) algorithm to ensure numerical stability and robust generalization. To improve interpretability, Local Interpretable Model-agnostic Explanations (LIME) are employed to elucidate how individual features influence model predictions. The results highlight the potential of explainable probabilistic learning for proactive identification of lost-circulation risks, optimized design of lost circulation materials (LCM), and reduction of operational uncertainties in drilling applications.

## I. Introduction (Heading 1)

Lost circulation is one of the most critical challenges in drilling operations, occurring when drilling fluid is partially or completely lost into the formation. The severity of losses ranges from seepage (1–10 bbl/hr) to partial (10–500 bbl/hr) and complete losses (>500 bbl/hr) [1,2]. Such events can lead to severe complications including stuck pipe, blowouts, wellbore instability, poor cementing, reservoir damage, and in extreme cases, suspension of wells for years, with lost circulation accounting for up to 20% of total drilling costs [3, 4]. These problems are particularly prevalent in formations with narrow mud-weight windows, such as fractured or pressure-depleted zones, where contributing factors include wellbore geometry, mud rheology, formation permeability, in-situ stress, pump pressure, and flow rate [5]. Current mitigation strategies largely rely on preventive and remedial methods, with the addition of lost circulation materials (LCM) being the most common. These materials are intended to bridge or seal fractures, though their success remains inconsistent due to the limited understanding of the loss mechanisms. A key factor is the optimization of LCM particle size to match fracture width, which is difficult to predict because of numerous interacting parameters [6]. Estimating fracture aperture through fluid-loss rate models offers a path toward proactive design of LCM formulations, allowing tailored solutions that minimize excessive fluid losses.

In parallel, machine learning (ML) has gained significant traction in the petroleum drilling sector for handling nonlinear, complex, and data-intensive problems. A recent systematic review of 369 studies published between 2017 and 2023 highlights the growing role of artificial intelligence (AI) in lost circulation research, including prediction, detection, and mitigation. The review underscores AI's potential to reduce drilling costs, waste, and environmental impact while identifying gaps such as the need for real-time monitoring integration and more robust field-deployable models [10]. Several recent studies have advanced ML applications for lost circulation prediction. In the Marun oil field, hybrid models combining multilayer perceptron (MLP) and least squares support vector machines (LSSVM) with optimization algorithms (GA, PSO, COA) demonstrated superior predictive accuracy, guiding improved LCM design and drilling decisions [11]. Wood et al. (2022) evaluated ML and deep learning (DL) algorithms using a large, imbalanced dataset from the Azadegan oil field (65,376 records, 20 wells). Despite challenges in predicting rare severe and complete losses (<0.5% of cases), ensemble tree-based models (Random Forest, Decision Tree, AdaBoost, KNN) consistently outperformed DL methods, with Random Forest achieving the highest accuracy and reliability [12]. Similarly, Liu et al. (2023) introduced a whale optimization algorithm–BiLSTM (WOA-BiLSTM) for mud loss prediction in Chinese oilfields. By applying wavelet filtering and feature selection via Spearman correlation, the optimized BiLSTM achieved remarkable performance (R² = 0.984, RMSE = 0.225), predicting mud loss up to 10 minutes in advance and enabling timely well-control interventions [13]. Jafarizadeh et al. (2023) proposed a CNN-based predictive framework, combining Savitzky–Golay preprocessing and NSGA-II feature selection from the Marun field data [14]. Their results showed that CNN outperformed hybrid MELM–LSSVM models in accuracy and generalizability, especially for high mud loss rates. More recently, ensemble machine learning approaches have been developed to further improve prediction performance for drilling fluid losses in the Marun oil field [15, 16]. Collectively, these works highlight that robust ML and hybrid AI models can provide early warning, improve LCM design, reduce non-productive time, and enhance decision-making in drilling operations.

Despite these advances, the existing approaches often lack the ability to explicitly quantify prediction uncertainty and may struggle to capture complex nonlinear relationships inherent in the drilling data. To overcome these limitations, in this study, an explainable Gaussian process regression (GPR) framework is developed to predict the drilling fluid loss of circulation in the Marun oil field. The proposed approach combines the probabilistic strength of GPR with model-agnostic explainable AI techniques to provide both accurate and interpretable predictions. The GPR model captures the nonlinear relationships between drilling parameters and mud loss severity while quantifying predictive uncertainty, which is critical for decision-making under risk. To further enhance transparency and trust in the model, Local Interpretable Model-Agnostic Explanations (LIME) are employed to identify the key factors having notable influence on the GPR predictions. This integration of predictive modeling and explainability offers a practical framework for proactive well control and data-driven drilling optimization. The rest of the manuscript is organized as follows. Section II provides the details of the Marun oil field data. The proposed predictive modelling framework is discussed in Section III. The proposed modelling algorithm is applied to the Marun oil field data and results are discussed in Section IV. Finally, the paper gets concluded in Section V.

## II. Data collection from Marun oil field

Marun oil field is located in the southern part of the northern Dezful embayment, roughly at 30°51′13″ N and 49°50′19″ E. It lies along the northwest–southeast trend of the Zagros fold-and-thrust belt in southwestern Iran. Structurally, Marun is an asymmetric anticline about 67 km long and 7 km wide, containing two main oil reservoirs (Asmari and Bangestan) and one gas reservoir, Khami [17, 18]. The Asmari formation, which serves as the main production zone, is composed of interbedded carbonate, shale, and sandstone layers. In contrast, the Bangestan and Khami formations are mainly carbonates with shale interbeds. Due to long-term tectonic activity, the field is heavily folded and fractured, which has caused significant drilling fluid losses over the years, particularly in the Asmari reservoir. Between 2002 and 2006, 305 wells were drilled in Marun, most of them targeting the Asmari formation, while only 17 reached the Bangestan and 4 the Khami reservoirs. These geological conditions make Marun a suitable site for studying lost circulation behavior and testing predictive models based on GPR and LIME.

The dataset was acquired from onshore drilling operations in the Marun oil field [11]. The reservoir interval is highly fractured, a structural characteristic that contributes to frequent mud loss incidents during drilling. The compiled dataset comprises 2,820 samples with 19 variables obtained from multiple sources, including final well completion reports, daily drilling logs, mud reports, and relevant literature. All variables are grouped into three categories: drilling operation parameters, formation properties, and drilling fluid characteristics. The target variable (mud loss severity) is derived from the daily drilling reports, where the lost circulation events are documented. The severity of each event was quantified using the difference between the pump flow rate and the measured return flow at the wellbore outlet, representing the dependent variable in the predictive modeling framework.

## III. The proposed predictive methdology

This section presents a detailed discussion of the development of the proposed predictive modelling framework.

GPR is a non-parametric kernel-based ML method, which is one of the popular methods for regression [19]. Gaussian processes (GPs) are nothing but the extension of multivariate Gaussian distributions to infinite dimensionality. As opposed to assuming a specific form for the underlying relation $f(x)$ between the input variables and the response variable, GPR describes $f(x)$ as a multivariate Gaussian distribution, which is fully defined by its mean function $m(x)$ and covariance matrix $K(x, x')$:

$$f(x) \sim GP(m(x), K(x, x')) + \varepsilon, \quad (1)$$

where $m(x) = E[f(x)]$,

$K(x, x') = E[(f(x) - m(x))(f(x') - m(x'))]$,

$\varepsilon$ is Gaussian distributed noise with zero mean and variance $\sigma^2$.

It is assumed that the training data are generated from a GP with zero mean. Accordingly, the prior distribution of the output observations is a Gaussian distribution defined below.

$$y \sim N(0, K(x, x')). \quad (2)$$

The covariance matrix $K(x, x')$ is defined as

$$K(x, x') = \begin{bmatrix} k(x_1, x_1) & k(x_1, x_2) & \cdots & k(x_1, x_n) \\ k(x_2, x_1) & k(x_2, x_2) & \cdots & k(x_2, x_n) \\ \vdots & \vdots & \ddots & \vdots \\ k(x_n, x_1) & k(x_n, x_2) & \cdots & k(x_n, x_n) \end{bmatrix}, \quad (3)$$

where $k(x, x')$ is a covariance function parameterized by a set of hyper-parameters.

For a given test data $x_*$, the objective of the GPR model is to predict $y_*$ using the training data. In order to do this, GP assumes that both the training output $y$ and the test output $y_*$ have joint distribution defined below.

$$\begin{bmatrix} y \\ y_* \end{bmatrix} \sim N\left(0, \begin{bmatrix} K(x, x') & k_* \\ k_*^T & k_{**} \end{bmatrix}\right), \quad (4)$$

where $k_* = [k(x_*, x_1), \dots, k(x_*, x_n)]^T$, $k_{**} = k(x_*, x_*)$.

The actual test output $y_*$ is estimated as the mean $E(y_*)$ of the posterior predictive distribution bestowed in Eq. (5).

$$P(y_*|X, y, x_*) = N(E(y_*), \sigma_*^2), \quad (5)$$

where $E(y_*) = k_*^T K^{-1} y$ and $\sigma_*^2 = k_{**} - k_*^T K^{-1} k_*$.

The variance of the posterior predictive distribution indicates uncertainty in the estimated output $\hat{y}_*$.

It can be noticed from the above discussion that the covariance function plays a key role in obtaining predictions from the GPR model. In the present work, exponential function is chosen as the covariance function:

$$k(x, x') = \sigma_f^2 \exp\left(-\frac{(x-x')^T(x-x')}{2\sigma_l^2}\right), \quad (6)$$

where $\sigma_l$ is the characteristic scale length, and $\sigma_f$ is the signal standard deviation.

The unknown parameters (parameters of Eq. (6) and noise standard deviation, $\sigma_n$) are determined by maximizing the log-likelihood of the training data:

$$J = \underset{\theta}{\arg\max}(\log P(y|X)) \quad (7)$$

$$J = \underset{\theta}{\arg\max}\left(-\frac{1}{2}y^T K^{-1} y - \frac{1}{2}\log|K| - \frac{N}{2}\log(2\pi)\right). \quad (8)$$

Eq. (8) is optimized using L-BGFS (limited memory Broyden–Fletcher–Goldfarb–Shanno) algorithm (Tipping and Faul 2003).

To improve the interpretability of the proposed GPR model, LIME is employed [20, 21]. This method provides quantitative insights into how the input features influence the model's predictions, enabling a clearer understanding of the complex relationships governing the drilling fluid loss. LIME explains a complex model by building a simpler, easy-to-understand model (surrogate model) around a specific prediction. It works locally, meaning it focuses on explaining one prediction at a time. The basic idea is as follows:

**Step 1**: Create $\{z_i\}_{i=1}^N$ by sampling near $x_k$ ($k = 1, ..., K$) (where $x_k$ is a given input sample from an independent test dataset and $K$ is the number of input samples in the validation dataset) by perturbing $x$ with the help of Gaussian or uniform noise.

**Step 2**: Compute $y_i = f(z_i)$, $f$ is the GPR model.

**Step 3**: Calculate the distance between the original input sample $x$ and each generated sample $z_i$.

$$\pi_x(z_i) = exp\left(\frac{D(x,z_i)^2}{\sigma^2}\right) \quad (9)$$

where $\sigma$ is the kernel width, small values indicate tight locality.

**Step 4**: Fit an interpretable surrogate model near $x_k$ by solving the following weighted, sparsity-encouraging regression to predict $y_i$ from $z_i$.

$$\hat{g}_x = arg \min_{g \in \mathcal{G}} \sum_{i=1}^N \pi_x(z_i)(y_i - g(z_i))^2 + \Omega(g) \quad (10)$$

$\mathcal{G}$ is a class of linear functions $g(z) = \beta_0 + \sum_{j=1}^d \beta_j z_j$, $\Omega(g)$ is a sparsity term (for instance L1 regularization).

The local coefficients are denoted by $\beta_x = (\beta_1, ..., \beta_d)$, where $d$ is the number of dimensions of $x_k$. The local importance of feature $j$ around $x_k$ is given by $|\beta_{xj}|$ and the sign of $\beta_{xj}$ signifies direction.

**Step 5**: As LIME is local, a local surrogate model is built for each input sample in the validation dataset. Then aggregate the coefficients of the local models to attain global measures.

Let $\beta_j^{(k)}$ be the coefficient for feature $j$ in the $k$-th local model. The global influence of each input feature is calculated by considering one or all the following global scores.

Mean absolute contribution

$$S_j^{(mean|\cdot|)} = \frac{1}{K}\sum_{k=1}^K |\beta_j^{(k)}| \quad (11)$$

Support frequency

$$S_j^{(freq)} = \frac{1}{K}\sum_{k=1}^K 1\{\beta_j^{(k)} \neq 0\} \quad (12)$$

Weighted average

$$S_j^{(w-mean|\cdot|)} = \frac{\sum_{k=1}^K w_k|\beta_j^{(k)}|}{\sum_{k=1}^K w_k}, w_k = \max(R_k^2, 0) \quad (13)$$

where $R_k^2$ is the coefficient of determination of the $k$-th local surrogate model.

**Step 6**: The global scores can be used to select features that have significant impact on the model predictions. A subset of features can be selected in the three approaches explained below.

- Elbow on the cumulative contribution curve of $S_j^{(w-mean|\cdot|)}$
- Pick features that appear in, say, ≥ 60-80% of bootstrapped runs
- Evaluate model performance as you include features and stop when the model performance ceases to stop.

Retrain the GPR model on the selected features only.

## IV. RESULTS AND DISCUSSIONS

Table I presents the physical descriptions of the input and output variables used in the Marun oil field dataset. The corresponding descriptive statistics are summarized in Tables S1 and S2 of the Supplementary Material, which can be accessed through the first author's GitHub repository (https://github.com/seshu-damarla/Explainable-GPR-for-Predicting-Drilling-Fluid-Loss-of-Circulation/tree/main).

Prior to the model development, data preprocessing was conducted to ensure the quality and usability of the dataset. Duplicate samples were identified and removed. As the raw measurements exhibited significant noise (see Figures S1 and S2 in the Supplementary Material), a Savitzky–Golay filter was applied for denoising. The resulting smoothed features are shown in Figures S3 and S4 of the Supplementary Material.

TABLE I. DETAILS OF MARUN OIL DATASET

| Variable | Symbol | Unit |
|---|---|---|
| Northing | $X_1$ | M |
| Easting | $X_2$ | M |
| Depth | $X_3$ | M |
| Meterage | $X_4$ | M |
| Drilling time | $X_5$ | Hr |
| Formation type | $X_6$ | – |
| Hole size | $X_7$ | In |
| Weight on bit | $X_8$ | 1000 lb |
| Flow rate | $X_9$ | Gpm |
| Mud weight | $X_{10}$ | Pcf |
| Marsh funnel viscosity | $X_{11}$ | – |
| Retort solid | $X_{12}$ | % |
| Pore pressure | $X_{13}$ | Psi |
| Fracture pressure | $X_{14}$ | Psi |
| $FAN_{600}/FAN_{300}$ | $X_{15}$ | – |
| $Gel_{10}$ min/$Gel_{10}$ s | $X_{16}$ | – |
| Pump pressure | $X_{17}$ | Psi |
| Bit rotational speed | $X_{18}$ | RPM |
| Mud-loss severity | Y | bbl/hr |

The relationships between the input variables and the output variable are inherently nonlinear, as illustrated in Figure S5. Consequently, conventional linear models fail to achieve satisfactory predictive performance, justifying the proposed predictive modelling approach The dataset was partitioned into training and test sets. The training set was used to determine the optimal values of the hyperparameters of the GPR model (Table II) by using the L-BFGS optimization technique. The test dataset was employed to evaluate the prediction performance of the GPR model.

TABLE II. OPTIMUM VALUES OF HYPERPARAMETERS

| Hyper Parameter | Optimum Value | Hyper Parameter | Optimum Value |
|---|---|---|---|
| $\sigma_f$ | 1.07 | $\sigma_{l9}$ | 1.61 |
| $\sigma_n$ | 0.00059 | $\sigma_{l10}$ | 453 |
| $\sigma_{l1}$ | 2.1 | $\sigma_{l11}$ | 0.595 |
| $\sigma_{l2}$ | 2.01 | $\sigma_{l12}$ | 0.812 |
| $\sigma_{l3}$ | 0.734 | $\sigma_{l13}$ | 0.435 |
| $\sigma_{l4}$ | 2.3 | $\sigma_{l14}$ | 1.13 |
| $\sigma_{l5}$ | 7.54 | $\sigma_{l15}$ | 0.873 |
| $\sigma_{l6}$ | 1.66 | $\sigma_{l16}$ | 0.74 |
| $\sigma_{l7}$ | 2.04e+03 | $\sigma_{l17}$ | 2.98 |
| $\sigma_{l8}$ | 3.09e+03 | $\sigma_{l18}$ | 2.11 |

The predictions obtained from the GPR model exhibit strong agreement with the actual mud loss severity values, as shown in Figures 1 and 2. Figure 3 illustrates the predictive uncertainty associated with each GPR estimate, while Figure 4 provides a closer examination of the first 150 test samples. The results demonstrate that the model yields lower uncertainty for test samples that are like the training data, whereas the uncertainty widens for samples that deviate from the training distribution. This behavior reflects the intrinsic property of GPR, where the predictive variance increases with the distance from known data points in the feature space.

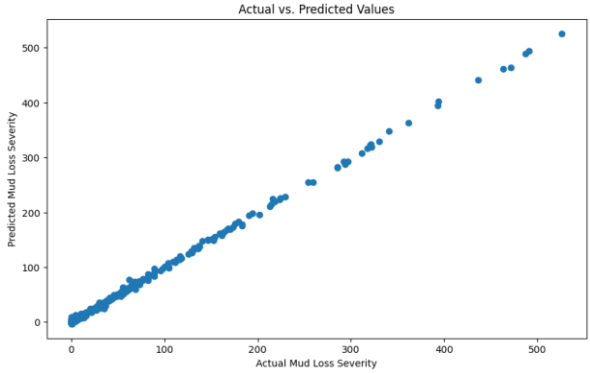

Fig. 1. Comparison of actual and GPR predictions on test dataset

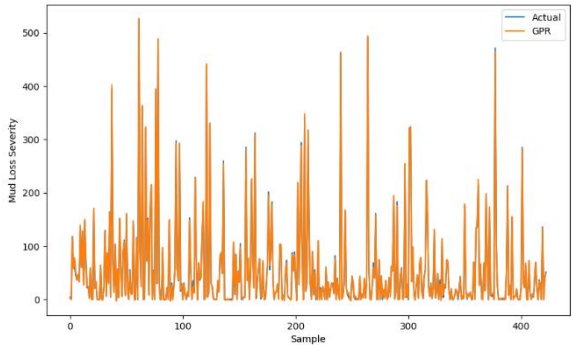

Fig. 2. Variation of actual and GPR-predicted mud-loss severity with sample index

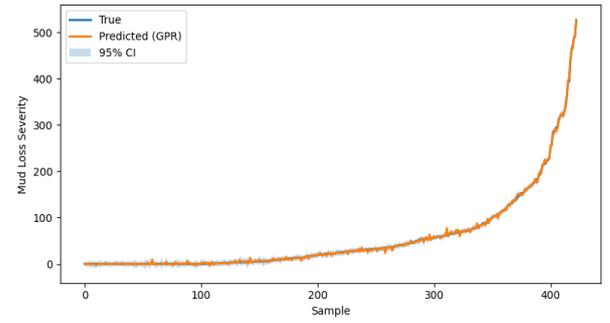

Fig. 3. GPR predictions with 95% confidence band.

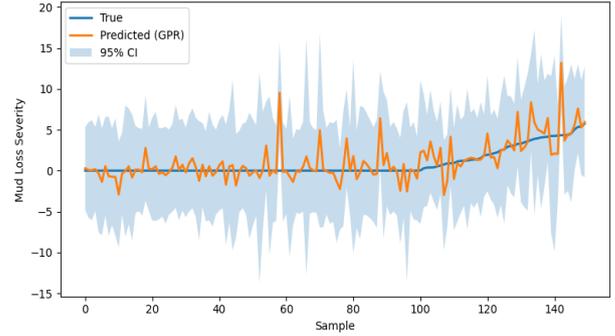

Fig. 4. Closer look at GPR predictions for the first 150 test samples.

TABLE III. GLOBAL IMORTANCE SCORES OBTAINED FROM LIME

| Feature | Absolute mean value | Actual mean value |
|---|---|---|
| $X_{12}$ | 1.460513 | 1.460513 |
| $X_{11}$ | 0.949101 | 0.948362 |
| $X_2$ | 0.853698 | 0.849887 |
| $X_{13}$ | 0.811951 | 0.797202 |
| $X_{17}$ | 0.644782 | 0.615994 |
| $X_1$ | 0.614017 | 0.178815 |
| $X_{14}$ | 0.458400 | 0.105129 |
| $X_{16}$ | 0.378474 | 0.066320 |
| $X_3$ | 0.349090 | 0.056977 |
| $X_{15}$ | 0.328884 | 0.038983 |
| $X_5$ | 0.318241 | -0.039515 |
| $X_{10}$ | 0.316163 | -0.055523 |
| $X_4$ | 0.312646 | -0.093871 |
| $X_7$ | 0.309683 | -0.129766 |
| $X_8$ | 0.307406 | -0.244505 |
| $X_9$ | 0.298239 | -0.330829 |
| $X_6$ | 0.288580 | -0.396782 |
| $X_{18}$ | 0.249394 | -0.588353 |

To quantify the influence of individual input variables on the model predictions, the LIME technique was applied to the developed GPR model. Table III summarizes both the absolute and actual mean LIME weights for all features. The absolute mean values represent the overall strength of each feature's contribution to the predicted mud loss, while the actual mean values indicate the direction of influence i.e. the positive values correspond to features that increase predicted mud loss, whereas the negative values correspond to features that decrease it. From Table III, features $X_{12}$, $X_{11}$, $X_2$, $X_{13}$, and $X_{17}$ exhibit the highest absolute mean weights, signifying their dominant contribution to model predictions. In particular, $X_{12}$ (1.46) and $X_{11}$ (0.95) show substantially higher importance compared to the remaining features, highlighting their strong predictive relevance. Features $X_1$, $X_{14}$, and $X_{16}$ provide moderate contributions, while features $X_5$ through $X_{18}$ demonstrate relatively small absolute mean weights, implying limited influence on the prediction outcomes.

The actual mean weights further reveal that the top-ranked features generally have a positive effect on the predicted mud loss, whereas features $X_5$, $X_{10}$, $X_4$, $X_7$, $X_8$, $X_9$, $X_6$, and $X_{18}$ exhibit negative contributions, suggesting their inverse relationship with the mud loss severity. These patterns align with physical expectations in drilling operations, where certain operational or formation parameters act as driving factors for fluid loss, while others have mitigating effects. Overall, the LIME-based analysis provides clear interpretability of the GPR model by identifying the most influential variables and their direction of impact. This insight supports feature selection for model simplification and enhances the understanding of how individual inputs govern the drilling fluid loss behavior.

## V. Conclusions

This study developed an explainable probabilistic machine learning framework using Gaussian process regression (GPR) for predicting drilling fluid loss of circulation in the Marun oil field. The model successfully captured the complex nonlinear interactions among the drilling, formation, and the mud parameters while quantifying predictive uncertainty, which is an essential capability for high-risk decision-making in drilling operations. The probabilistic nature of the GPR model enabled reliable prediction intervals, reflecting higher uncertainty for unfamiliar operating conditions and smaller variance within the data domain of the training set. Explainable artificial intelligence technique was incorporated to enhance interpretability. LIME was applied to quantify the contribution of each input feature to the model's predictions on the test dataset. The global importance scores derived from the aggregated absolute mean LIME weights provided valuable insights into the relative influence of each variable. The signed mean weights revealed the direction of influence whether a feature contributed to increasing or decreasing mud-loss severity thereby distinguishing between causative and mitigating factors. Features such as flow rate, pore pressure, pump pressure, and mud rheology indices consistently exhibited high importance, confirming their dominant role in the mud-loss behavior. The integration of the GPR model with LIME-based explainability presents a powerful and transparent framework for the mud-loss prediction. The findings highlight the complementary strengths of probabilistic learning and interpretable AI: accurate predictions with quantified confidence and actionable understanding of feature influence.